\documentclass[lettersize,journal]{IEEEtran}
\usepackage{amsmath,amsfonts}
\usepackage{algorithmic}
\usepackage{array}
\usepackage{textcomp}
\usepackage{stfloats}
\usepackage{url}
\usepackage{verbatim}
\usepackage{graphicx}
\def\BibTeX{{\rm B\kern-.05em{\sc i\kern-.025em b}\kern-.08em
    T\kern-.1667em\lower.7ex\hbox{E}\kern-.125emX}}
\usepackage{balance}
\usepackage{times}
\usepackage{latexsym}
\usepackage[T1]{fontenc}
\usepackage[utf8]{inputenc}
\usepackage{microtype}
\usepackage{inconsolata}
\usepackage{subfigure}
\usepackage{graphicx}
\usepackage{multirow}
\usepackage{algorithmic}
\usepackage{algorithm}
\usepackage{amsmath,amsfonts}
\usepackage{textcomp}
\usepackage{stfloats}
\usepackage{url}
\usepackage{verbatim}
\usepackage{balance}
\usepackage{array}
\usepackage{amssymb}
\usepackage{booktabs}
\usepackage{xcolor}  
\usepackage{color}
\usepackage{colortbl}

\definecolor{label_color}{RGB}{255, 204, 153}
\definecolor{text_color}{RGB}{205, 235, 139}
\definecolor{output_color}{RGB}{224, 213, 231}
\definecolor{sememe_color}{RGB}{255, 255, 136}

\definecolor{uc}{RGB}{77, 191, 249}
\definecolor{bc}{RGB}{5,178,83}

\definecolor{hdash}{RGB}{191, 191, 191}

\definecolor{backgroud}{RGB}{240,240,240}

\definecolor{nmgray}{RGB}{229,229,229}
\usepackage{CJKutf8}

\usepackage{arydshln}

\definecolor{ljcolor}{RGB}{34,139,34}

\begin{document}
\title{\textit{TKDP}: \underline{T}hreefold \underline{K}nowledge-enriched \underline{D}eep \underline{P}rompt Tuning for Few-shot Named Entity Recognition}
\author{Jiang Liu, Hao Fei, Fei Li, Jingye Li, Bobo Li, Liang Zhao, Chong Teng and Donghong Ji
\thanks{Jiang Liu, Fei Li, Jingye Li, Bobo Li, Chong Teng, Donghong Ji are with the Key Laboratory of Aerospace Information Security and Trusted Computing, Ministry of Education, School of Cyber Science and Engineering, Wuhan University (e-mail: liujiang@whu.edu.cn; foxlf823@gmail.com; theodorelee@whu.edu.cn; 
boboli@whu.edu.cn; 
tengchong@whu.edu.cn; dhji@whu.edu.cn).
}
\thanks{Hao Fei is with the School of Computing, National University of Singapore (e-mail: haofei37@nus.edu.sg).}
\thanks{Liang Zhao is with the University of São Paulo, Brazil (e-mail: zhao@usp.br).}
}

\markboth{Journal of \LaTeX\ Class Files,~Vol.~18, No.~9, September~2020}%
{How to Use the IEEEtran \LaTeX \ Templates}

\maketitle

\begin{abstract}
Few-shot named entity recognition (NER) exploits limited annotated instances to identify named mentions. 
Effectively transferring the internal or external resources thus becomes the key to few-shot NER.
While the existing prompt tuning methods have shown remarkable few-shot performances, they still fail to make full use of knowledge.
In this work, we investigate the integration of rich knowledge to prompt tuning for stronger few-shot NER.
We propose incorporating the deep prompt tuning framework with threefold knowledge (namely \emph{TKDP}), including the internal 1) \emph{context knowledge} and the external 2) \emph{label knowledge} \& 3) \emph{sememe knowledge}.
TKDP encodes the three feature sources and incorporates them into the soft prompt embeddings, which are further injected into an existing pre-trained language model to facilitate predictions.
On five benchmark datasets, our knowledge-enriched model boosts by at most 11.53\% F1 over the raw deep prompt method, and significantly outperforms 8 strong-performing baseline systems in 5-/10-/20-shot settings, showing great potential in few-shot NER.
Our TKDP can be broadly adapted to other few-shot tasks without effort.
\end{abstract}


\section{Introduction}
Named entity recognition aims to extract named mentions (e.g., people, organizations and locations) from text \cite{nerdef1,ner_def2}.
As one of the fundamental tasks in natural language processing (NLP), NER often serves as an upstream component of more complex tasks such as information retrieval \cite{information_retrieval}, relation extraction \cite{relation_extraction} and machine reading comprehension \cite{mrc} etc. 
Within the last decade, NER has achieved remarkable success with the aid of deep learning techniques, relying on large-scale standard corpora \cite{lc_paradigm,mrc_paradigm}.
However, manually constructing annotations is time-consuming, labor-intensive and even infeasible, especially for certain fields, e.g., medical \cite{prototype}.
This thus demands the research of few-shot NER, to learn a NER system with few labeled examples \cite{struct}.

\begin{figure}[!t]
\centering
\includegraphics[width=1\columnwidth]{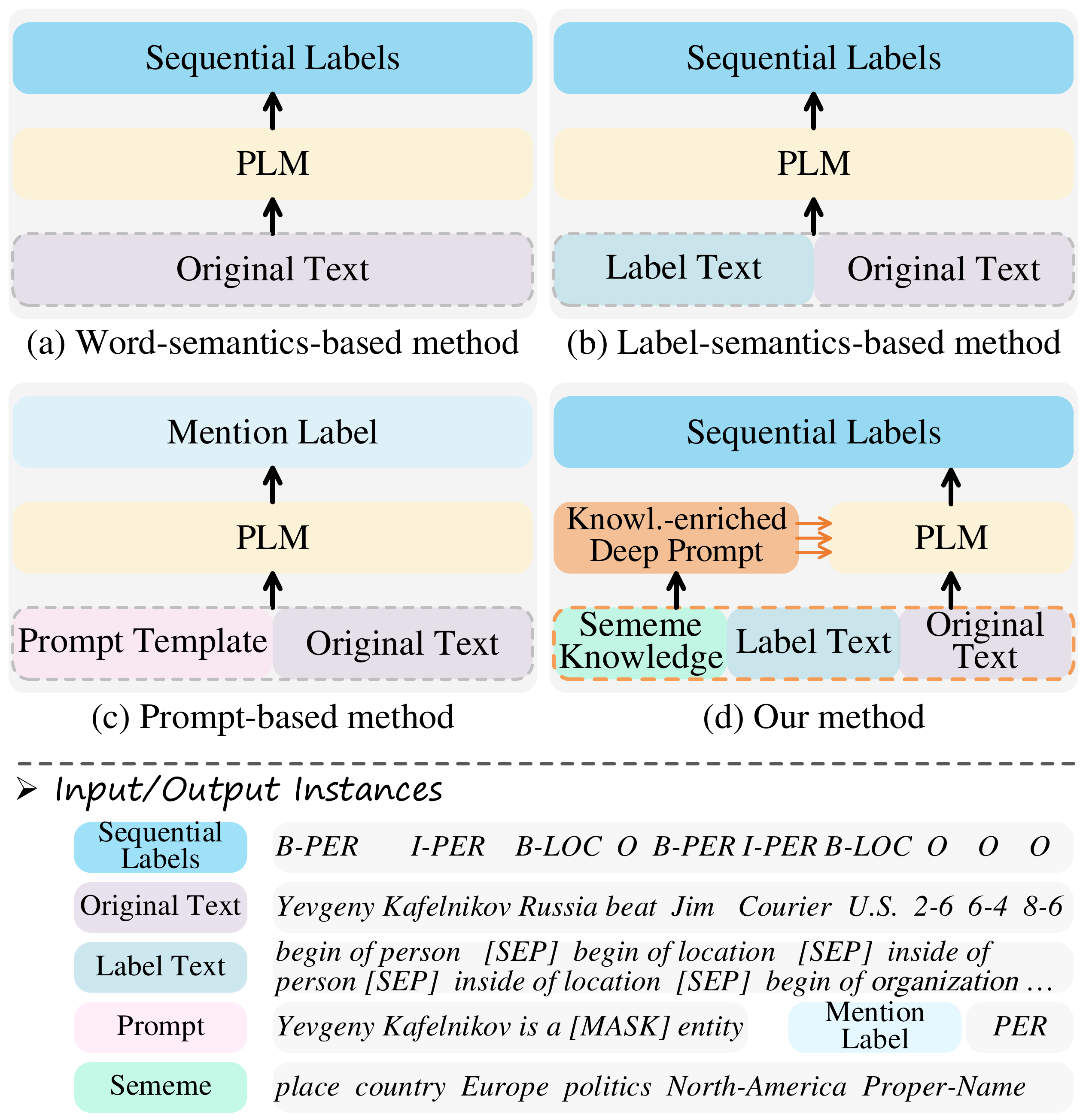}
\caption{Comparison between existing few-shot NER methods and our knowledge-enriched deep prompt based framework that makes use the threefold knowledge features: sememe, label and context knowledge. 
}
\label{fig1} 
\end{figure}

Existing few-shot NER works have made considerable progress \cite{entity-base,seefew}, which can be technically grouped into three paradigms: \textbf{word-semantic-based method}, \textbf{label-semantic-based method} and \textbf{prompt-based method}. 
As shown in Figure \ref{fig1}, the word-semantic-based methods \cite{struct} depend solely on the input words and their context, while the label-semantic-based methods \cite{prototype} additionally make use of the label knowledge.
In contrast, the prompt-based methods \cite{template} are built upon the current pre-trained language models (PLM) \cite{PLM_BERT,PLM_roberta,PLM_bart}, trying to guide the model to identify entities with pre-built natural language templates, i.e., prompt texts.

While prompt-based methods have achieved better few-shot performances \cite{prompt_text_classfiy,prompt_}, the construction of hard discrete prompt templates can be much experience oriented and cause unstable results.
Also for the NER task, enumerating all possible spans to build prompts are low-efficient.
The recent progress of deep prompt tuning \cite{ptuingv2} helps relieve these concerns by considering the soft and continuous prompts, which helps few-shot NER achieve the remarkable result \cite{lightner}.
Notwithstanding, we note that there are still sufficient rooms to improve.
Prior studies extensively reveal that NER relies much on semantic understanding, where the knowledge (internal and external) plays the critical role, especially for the few-shot scenario \cite{seyler-etal-2018-study,sui-etal-2019-leverage}, while we observe the current prompt-based work scarcely considers this.
Therefore, this paper investigates the feasibility of incorporating rich knowledge for better prompt-based few-shot NER.

In fact, the deep prompt method is effective in learning semantics from the internal \textbf{context features}, yet the other external information is overlooked, such as the \textbf{label texts} and the \textbf{sememe knowledge}.
Essentially, the entity labels provide rich semantic clues to describe what an entity mention would be \cite{chen-etal-2022-shot}, while the sememe \cite{sememe,sememe_BabelNet} is a finite and semantically indivisible set of words that can compose the meaning of a word and extend the word semantics.
In this work, we take the initiative to enhance the deep prompt method with all these three types of internal and external knowledge, aka, threefold knowledge-enriched deep prompt tuning (namely \textit{TKDP}).
We depict the differences between our method and previous methods in Figure \ref{fig1}.

\begin{figure*}[!t]
\centering
\includegraphics[width=1\textwidth]{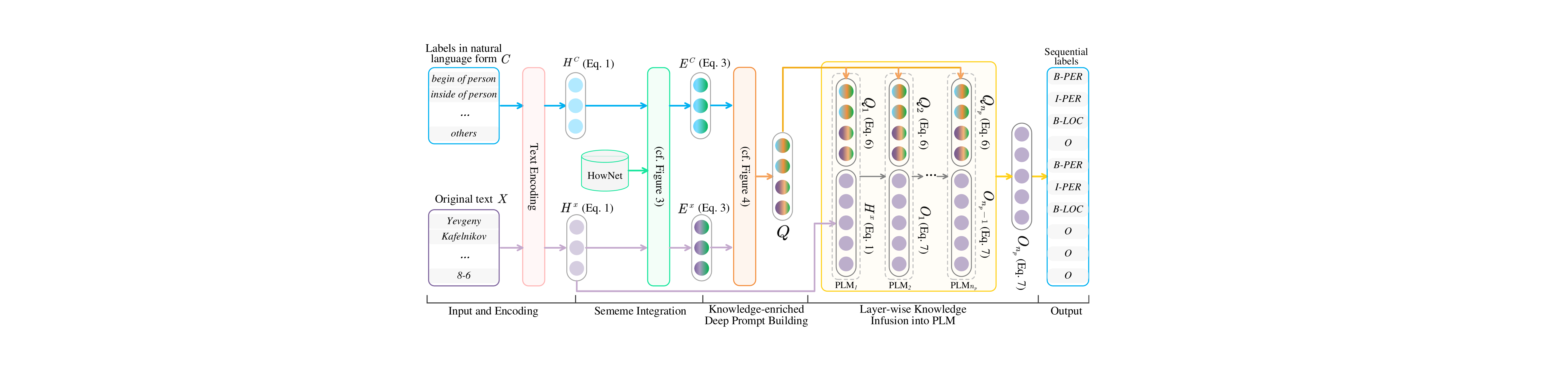}
\caption{The overall architecture of our {\textbf{TKDP}} model. $\boldsymbol{H}^{x}$ and $\boldsymbol{H}^{C}$ are text embeddings and label embeddings, respectively. 
$\boldsymbol{E}^{x}$ and $\boldsymbol{E}^{C}$ represent sememe-enhanced text embeddings and label embeddings. 
$\boldsymbol{Q}_{i}$ is the knowledge-enhanced prompt embedding, which is inserted into $i$-th layer of PLM.  
} 
\label{model_frame} 
\end{figure*}

As shown in Figure \ref{model_frame}, our framework is built based on an existing PLM, equipped with knowledge-rendered soft prompts.
Specifically, TKDP first takes as inputs the label descriptions and sentential words.
We then embed the sememe representations from HowNet into the prior label and text contextual representations for semantics enhancement.
Afterward, based on the above sememe-enhanced label and text representations, we generate two types of knowledge-enriched deep prompt embeddings, which are then fused into different layers of PLM for deep interaction.
With the incorporation of sememe, label and text context knowledge, the framework is expected to output \emph{BIO} label sequences more accurately.

Extensive experiments are conducted on total five public benchmark datasets across multiple domains.
The results show that our method significantly outperforms 8 strong baseline systems in all three few-shot settings (5-/10-/20-shots);
boosting the results by at most 11.53\% F1 over the raw soft prompt method without any knowledge, demonstrating the urgent necessity to incorporate the proposed three kinds of knowledge.
We further show that our system wins the current popular prompt learning systems on the task by large margins, including the prompt-tuning \cite{prompt}, p-tuning \cite{ptuning}, prefix-tuning \cite{prefix} and discrete template-based prompt methods \cite{entlmner,template}.

Our knowledge-enriched prompt tuning framework can be broadly applied to other few-shot NLP tasks in general purpose without much effort.
All the resources of this paper will be publicly available to facilitate related research.

\section{Related Work}
\subsection{Few-shot NER}
In this section, we briefly survey the existing few-shot NER methods under three categories.

\textbf{Word-semantic-based Methods} 
Earlier studies
learn the context semantics for few-shot NER from only the input words, i.e., inferring the entity words from their contexts.
For example, Yang and Katiyar (2020) \cite{struct} propose a model based on nearest neighbor learning and structured reasoning, where each token is represented in the tokenized examples of the support set by a contextual representation in the sentence. 
Das et al. (2022) \cite{container} use contrastive learning to infer the distribution distance of its Gaussian embedding, thus to reduce the distance of token embedding of similar entities and increase the distance of token embedding of different entities. 
One of the key issues of this type of method is the feature deficiency; with only word semantics itself, the NER model is hard to learn a robust inductive bias for recognizing unseen named entities.

\textbf{Label-semantic-based Methods} 
For few-shot NER, the context features for models to learn from labeled data are limited, and thus label information is considered to be applied in label-semantic-based methods.
Intuitively, mention labels describe the attributes of specific entity prototypes in natural language, offering rich extended semantic features of mentions for few-shot NER \cite{chen-etal-2022-shot}.
Hou et al. (2020) \cite{slot} first leverage the semantics of label names for few-shot NER. 
Huang et al. (2021) \cite{prototype} and Ji et al. (2022) \cite{entity-base} create prototype representations from different entity types, and assign labels through the nearest neighbor criterion according to the label dependencies.
Ma et al. (2022) \cite{labelbert} proposed a simple two-tower model to incorporate the additional signals from label semantic information.

\textbf{Prompt-based Methods} 
In recent years, the birth of large PLMs \cite{gpt3} has activated the generative paradigm.
Based on PLMs, prompt tuning techniques have shown prominent few-shot performances for multiple NLP tasks \cite{prompt_text_classfiy,prompt_}, yet most of the prompt templates are customized for sentence-level classification rather than NER.
Later,  Cui et al. (2021) \cite{template} first apply the prompt tuning to the sequence tagging task, which create prompt templates by enumerating all possible mention spans.
Lee et al. (2022) \cite{Demonstration} and Ma et al. (2022) \cite{entlmner} follow the same practice of the hard prompt tuning for few-shot NER.
As cast earlier, building templates are experience-oriented, and it is also low-efficient to enumerate the spans.
Instead of constructing templates for discrete prompts, recently,  Chen et al. (2022) \cite{lightner} leverage the deep prompt tuning \cite{ptuingv2} for few-shot NER, where the prompts are soft and continuous representations.
Although achieving remarkable results, current deep prompt methods still overlook the full utilization of semantic features, which thus motivates our work of knowledge-enriched deep prompt tuning. 

\begin{figure*}[!t]
\centering
\includegraphics[width=0.75\textwidth]{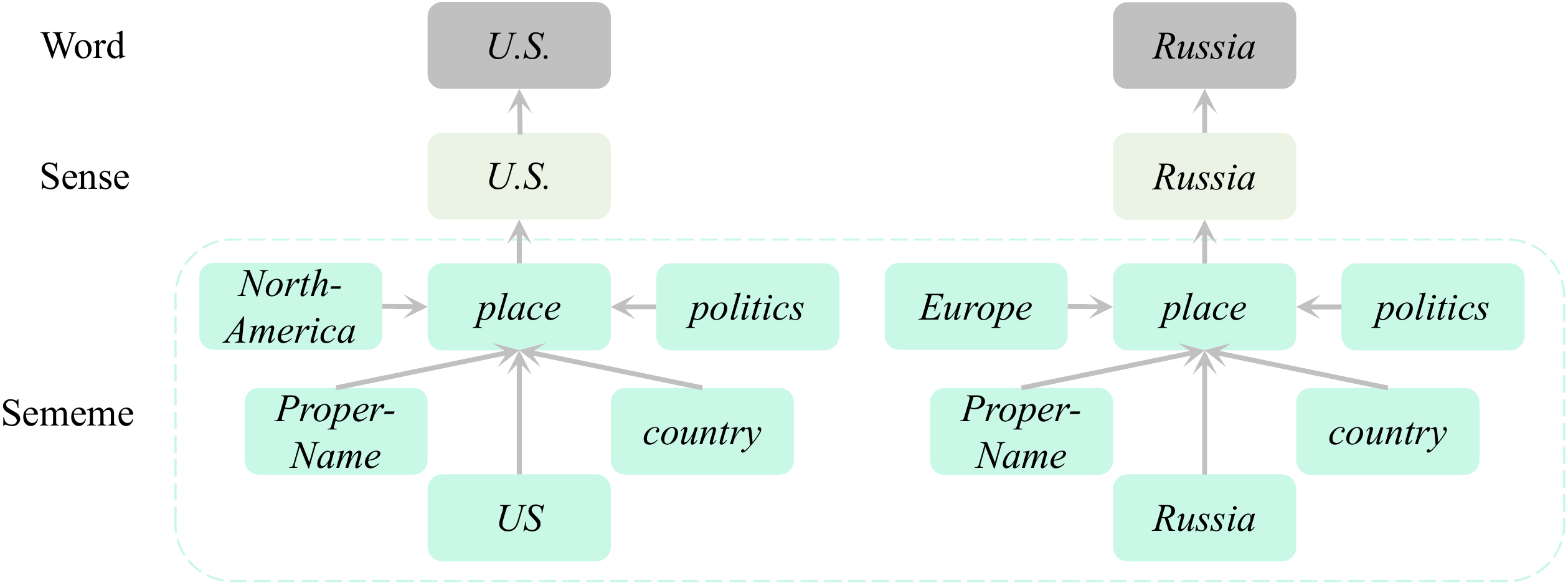}
\caption{Illustration of word, sense and sememe in HowNet.} 
\label{word_sense_sememe} 
\end{figure*}

\begin{table*}[h]
\caption{The natural language form corresponding to the original entity type of all datasets.
\label{tab:Natural_language_appendix}}
\fontsize{9.5}{9.5}\selectfont
\setlength{\tabcolsep}{0.8mm}
\centering
\resizebox{1\textwidth}{!}{
\begin{tabular}{lllclll}
\cellcolor{backgroud}{{\bf{Datasets}}}&\cellcolor{backgroud}{{\bf{Original Label Type}}}&\cellcolor{backgroud}{{\bf{Natural Language Form}}}&\cellcolor{backgroud}{\phantom{}}&\cellcolor{backgroud}{{\bf{Datasets}}}&\cellcolor{backgroud}{{\bf{Original Label Type}}}&\cellcolor{backgroud}{{\bf{Natural Language Form}}}\\
\multirow{4}{*}{\bf{CoNLL-2003}}&ORG&organization&\cellcolor{backgroud}{\phantom{}}&\multirow{12}{*}{\bf{MIT-Movie}}&Actor	&actor\\
{}&MISC&miscellaneous&\cellcolor{backgroud}{\phantom{}}&{}&Plot&plot\\
{}&PER&person&\cellcolor{backgroud}{\phantom{}}&{}&Opinion&opinion\\
{}&LOC&location&\cellcolor{backgroud}{\phantom{}}&{}&Award&award\\
\cline{0-2}
\multirow{6}{*}{\bf{WNUT-2017}}&location&location&\cellcolor{backgroud}{\phantom{}}&{}&Year&year\\
{}&group&group&\cellcolor{backgroud}{\phantom{}}&{}&Genre&genre\\
{}&corporation&corporation&\cellcolor{backgroud}{\phantom{}}&{}&Origin&origin\\
{}&person&person&\cellcolor{backgroud}{\phantom{}}&{}&Director&director\\
{}&creative\_work&creative work&\cellcolor{backgroud}{\phantom{}}&{}&Soundtrack&soundtrack\\
{}&product&product&\cellcolor{backgroud}{\phantom{}}&{}&Relationship&relationship\\
\cline{0-2}
\multirow{8}{*}{\bf{MIT-Restaurant}}&Rating&rating&\cellcolor{backgroud}{\phantom{}}&{}&Character\_Name&character name\\
{}&Amenity&amenity&\cellcolor{backgroud}{\phantom{}}&{}&Quote&quote\\
\cline{4-7}
{}&Location&location&\cellcolor{backgroud}{\phantom{}}&\multirow{5}{*}{\bf{JNLPBA}}&DNA&deoxyribonucleic acid\\
{}&Restaurant\_Name&restaurant name&\cellcolor{backgroud}{\phantom{}}&{}&protein&protein\\
{}&Price&price&\cellcolor{backgroud}{\phantom{}}&{}&cell\_type&cell type\\
{}&Hours&hour&\cellcolor{backgroud}{\phantom{}}&{}&cell\_line&cell line\\
{}&Dish&dish&\cellcolor{backgroud}{\phantom{}}&{}&RNA&ribonucleic acid\\
\cline{4-7}
{}&Cuisine&cuisine&\cellcolor{backgroud}{\phantom{}}&{}&{}&{}\\
\cline{0-2}
\end{tabular}
}
\end{table*}

\subsection{Sememe}
\textbf{HowNet and Sememes}
HowNet is one of the most famous term knowledge bases, which defined more than 100,000 Chinese and English words with 2187 terms \cite{openhownet}.  It describes words or phrases through morphemes, which are the smallest units of semantic concepts \cite{hownet_appendix}. 
Recent studies have shown that HowNets can provide rich well-recognized semantic representations that facilitate downstream NLP tasks, such as word similarity computation \cite{sememe_word_similarity}, word sense disambiguation \cite{sememe_disambiguation}, event detection \cite{sememe_event}, word representation learning \cite{sememe_representation}, language modeling \cite{sememe_lang_model}, lexicon expansion \cite{sememe_lexicon}, relation extraction \cite{sememe_relation} and semantic
rationality evaluation \cite{sememe_sre}. 

Jin et al. (2018) \cite{sememe_example1} incorporate Chinese character information into their sememe prediction model and achieved a performance boost. Qi et al. (2018) \cite{sememe_cross} make the first attempt to use cross-lingual sememe prediction to build a sememe knowledge base for other languages. Qi et al. (2019) \cite{sememe_SC} incorporate sememe knowledge into the semantic composition model for the first time and verified its effectiveness. Qi et al. (2020) \cite{sememe_BabelNet} built a unified sememe knowledge base for multiple languages based on BabelNet (a multilingual encyclopedia dictionary). Lyu et al. (2021) \cite{sememe_Glyph} propose the glyph enhanced ghinese character representation, which optimizes the prediction of lexical sememes by focusing on finer-grained information. Qi et al. (2022) \cite{sememe_image} use image information for the prediction of lexical sememes for the first time.

\textbf{Sememes for Enriching Semantics}
Taking Figure \ref{word_sense_sememe} as example, the word \emph{"U.S."} can be represented as the combination of sememes:
\emph{"place"}, \emph{"politics"}, \emph{"North-America"}, \emph{"US"}, \emph{"Proper-Name"} and \emph{"country"}. That is to say, the sememe set completely describes the semantic space of a word and contains rich semantics. And this information is beneficial to the extraction of entities. For example, \emph{"U.S."} and \emph{"Russia"} share the latent sememe concept \emph{"place"}, which can strengthen their connection with the entity type \emph{"LOC"}.

\subsection{Knowledge-enhanced Prompt Tuning}
\label{relateknowledgeableord KP}

It is also common practice to leverage external or internal knowledge to enhance the downstream NLP tasks.
Here we briefly survey the literature concerning knowledge-enhanced prompt learning methods.
Shin et al. (2022) \cite{AutoPrompt} used a gradient guided search algorithm to automatically construct prompts for various task sets, consisting of raw input and trigger words (obtained by the gradient guided search algorithm), which are discrete prompts. The entire process does not involve fusion of knowledge, so it is different from our automatic construction of fusion knowledge prompts.
Hu et al. (2022) \cite{knowledgeable} incorporate the external knowledge into the verbalizer to form knowledgeable prompt tuning to improve and stabilize prompt tuning. 
Specifically, Hu et al. (2022) \cite{knowledgeable} use a knowledge base to extend the label word space of the verbalizer and use the PLM itself to refine the extended label word space before using the extended label word library for prediction. 
Chen et al. (2022) \cite{Knowprompt} inject the latent knowledge contained in relation labels into the construction of prompts with learnable virtual type words and answer words to solve the few-shot relation extraction problem.
Li et al. (2022) \cite{kp_event} set the event detection task as the condition generation task. 
Then, knowledge-injected prompts are constructed using external knowledge bases, and a prompt tuning strategy is leveraged to optimize the prompts.
This work differs from the above works in multiple aspects.
First, we consider the use of multiple knowledge from both internal and external sources.
Besides, we mainly consider the sememe knowledge from HowNet to enhance the text semantics, while the above works do not.
Finally, we focus on few-shot NER tasks, and these methods are more suitable for classification tasks and not for sequence labeling tasks. 

\section{Methodology}

Figure \ref{model_frame} presents our overall framework, which consists of five parts, including the input and encoding, the sememe integration layer, knowledge-enriched deep prompt construction, layer-wise knowledge infusion into PLM, and the output layer. 

\subsection{Input and Encoding}

Our system takes two input sources:
1) the original text $X=\{x_{1},x_{2},\ldots,x_{n} \}\in\mathbb{R}^{n}$ of length $n$, where $x_{i}$ represents the $i$-th word in the input text $X$; 
and 2) the natural language description of $L$ labels $C=\{C_{1},C_{2},\ldots,C_{L}\}\in\mathbb{R}^{L}$, where each label description $C_{j}=\{c_{j,1},c_{j,2},\ldots,c_{j,l} \}\in\mathbb{R}^{l}$ consists of $l$ description words.
For example, the description for the label $PER$ can be unfolded as:
\begin{equation}
\label{deqn_ex12a}
\begin{aligned}
B\mbox{-}PER &\to begin\enspace of\enspace person\nonumber \\
I\mbox{-}PER &\to inside\enspace of\enspace person\nonumber \\
O &\to others\nonumber
\end{aligned}
\end{equation}
The natural language forms corresponding to the original entity types of all datasets are detailed in Table \ref{tab:Natural_language_appendix}.
We encode input text and label description using the embedding layer of BERT \cite{PLM_BERT}. 
Then $X$ and $C$ are encoded as:
\begin{equation}
\label{deqn_ex2a}
\begin{aligned}
\boldsymbol{H}^{x}&=\{\boldsymbol{h}^{x}_{1},\boldsymbol{h}^{x}_{2},\ldots,\boldsymbol{h}^{x}_{n}\}\in\mathbb{R}^{n \times d_{h}} \,, \\
\boldsymbol{H}^{C}&=\{\boldsymbol{H}^{c}_{1},\boldsymbol{H}^{c}_{2},\ldots,\boldsymbol{H}^{c}_{L}\}\in\mathbb{R}^{L\times l \times d_{h}}\,,\\
where \enspace \boldsymbol{H}^{c}_{j}&=\{\boldsymbol{h}^{c}_{j,1},\boldsymbol{h}^{c}_{j,2},\ldots,\boldsymbol{h}^{c}_{j,l}\}\in\mathbb{R}^{l \times d_{h}}\,,
\end{aligned}
\end{equation}
where $\boldsymbol{h}^{x}_{i}$, $\boldsymbol{h}^{c}_{j,i}\in\mathbb{R}^{d_{h}}$ are the representations of the $i$-th word, and $ d_{h} $ represents the dimension of a word representation.

\begin{figure}[!t]
\centering
\includegraphics[width=1\columnwidth]{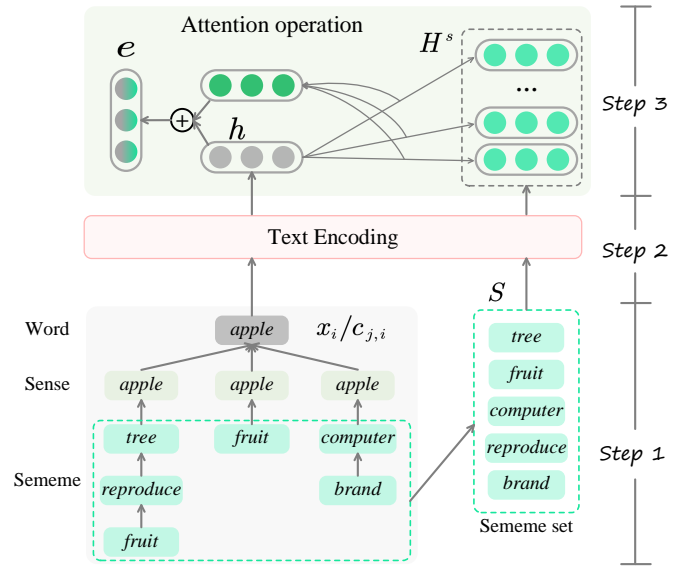}
\caption{
Illustration of the sememe knowledge integration.
\textbf{Step 1}, all the sememes (e.g., fruit) of the word (e.g., apple) are composed into the sememe set $S$.
\textbf{Step 2}, $S$ and the target word are encoded as representations, $\boldsymbol{H}^{s}$ and $\boldsymbol{h}$.
\textbf{Step 3}, $\boldsymbol{H}^{s}$ and $\boldsymbol{h}$ are fused via attention and added on $\boldsymbol{h}$ as residual.
\label{sememe_embedding_module} 
}
\end{figure}

\subsection{Sememe Integration}
\label{sec:Sememe}

Sememe is a language-independent smallest semantic unit \cite{sememe}, which has been leveraged as external knowledge base in many NLP tasks to enrich the semantics of inputs, e.g., word sense disambiguation \cite{sememe_disambiguation}, event detection \cite{sememe_event} and relation extraction \cite{sememe_relation}. 
In HowNet, a word is defined with one or more senses, and each sense contains one or more sememes. 
Now we consider embed the sememe information into our framework. 
As shown in Figure \ref{sememe_embedding_module}, first, we obtain the sememe set $S=\{s_{1},s_{2},\ldots,s_{m}\}\in\mathbb{R}^{m}$ of the target word from HowNet, where $m$ is the number of sememe, and then obtain the representation of the sememe set $\boldsymbol{H}^{s}=\{\boldsymbol{h}^{s}_{1},\boldsymbol{h}^{s}_{2},\ldots,\boldsymbol{h}^{s}_{m}\}\in\mathbb{R}^{m \times d_{h}}$,
where $\boldsymbol{h}^{s}_{i}\in\mathbb{R}^{d_{h}}$ represents the $i$-th sememe representation. 
Note that the embeddings for sememe words are shared with the input sentence and labels.


After obtaining the sememe representation, we use an attention mechanism to retrieve those sememes that are more related to the query word. 
The attention score of the $i$-th sememe representation $\boldsymbol{h}^{s}_{i}$ and the word representation $\boldsymbol{h}\in\mathbb{R}^{d_{h}}$\footnote{
Only in this subsection, for brevity we unify the previous word representations $\boldsymbol{h}^{x}_{i}$ and $\boldsymbol{h}^{c}_{j_{i}}$ to $\boldsymbol{h}$.
} is calculated by the euclidean distance:
\begin{equation}
\label{deqn_ex3a}
\begin{gathered}
r_{i}=\frac{d_{i}}{\sum_{i=1}^{m}d_{i}} \,, \enspace
d_{i}=\sqrt{\sum_{j=1}(h_{j}-h^{s}_{i,j}})^{2} \,,
\end{gathered}
\end{equation}
where $h_{j}$ and $h^{s}_{i,j}$ are the size of the $j$-th dimension of word representation and sememe representation respectively. Finally, the sememe-enhanced word representation $\boldsymbol{e}\in\mathbb{R}^{d_{h}}$ is obtained via $\boldsymbol{e}=\boldsymbol{h}+\sum_{i=1}^{m}r_{i}\cdot \boldsymbol{h^{s}_{i}}$.

Now we obtain the sememe-enriched text representation $\boldsymbol{E}^{x}\in\mathbb{R}^{n\times d_{h}}$ and the sememe-enriched label set representation $\boldsymbol{E}^{C}\in\mathbb{R}^{L\times l\times d_{h}}$:
\begin{equation}
\label{deqn_ex4a}
\begin{gathered}
\boldsymbol{E}^{x}=\{\boldsymbol{e}^{x}_{1},\boldsymbol{e}^{x}_{2},\ldots,\boldsymbol{e}^{x}_{n}\}\,, \\
\boldsymbol{E}^{C}=\{\boldsymbol{E}^{c}_{1},\boldsymbol{E}^{c}_{2},\ldots,\boldsymbol{E}^{c}_{L}\}\,,\\
\boldsymbol{E}^{c}_{j}=\{\boldsymbol{e}^{c}_{j,1},\boldsymbol{e}^{c}_{j,2},\ldots,\boldsymbol{e}^{c}_{j,l}\}\in\mathbb{R}^{l\times d_{h}}\,,
\end{gathered}
\end{equation}
where $\boldsymbol{e}^{x}_{i}$, $\boldsymbol{e}^{c}_{j,i}\in\mathbb{R}^{d_{h}}$ are the representations of the $i$-th word integrated into the sememe.

\subsection{Enriched Deep Prompt with Knowledge}

\begin{figure}[!t]
\centering
\includegraphics[width=1\columnwidth]{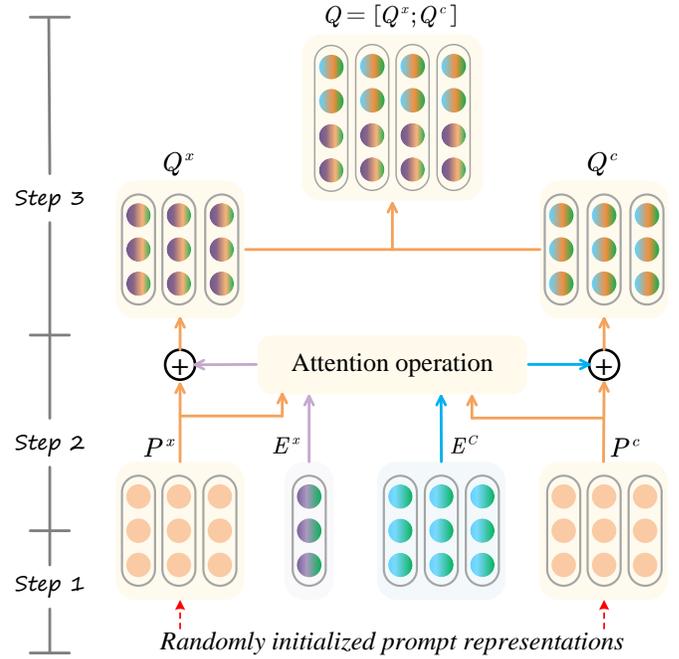}
\caption{
Building knowledge-enriched deep prompt representations.
\textbf{Step 1}, prompt embedding $\boldsymbol{P}^{x}$ and $\boldsymbol{P}^{c}$ are generated; 
\textbf{Step 2}, $\boldsymbol{P}^{x}$ and $\boldsymbol{P}^{c}$ are fused with the sememe-enhanced context representation $\boldsymbol{E}^{x}$ and label representation $\boldsymbol{E}^{C}$ via attention operation;
\textbf{Step 3}, the knowledge-enhanced context prompt $\boldsymbol{Q}^{x}$ and knowledge-enhanced label prompt $\boldsymbol{Q}^{c}$ are concatenated into unified deep prompt representation $\boldsymbol{Q}$.
} 
\label{KFM} 
\end{figure}

Next, we based on the $\boldsymbol{E}^{x}$ and $\boldsymbol{E}^{c}$ create the knowledge-enriched deep prompt representations.
As shown in Figure \ref{KFM}, we first use randomly initialize two prompt embeddings $\boldsymbol{P}^{x}$, $\boldsymbol{P}^{c}\in\mathbb{R}^{l_{p}\times n_{p} \times d_{h}}$, where $l_{p}$ represents the length of the prompt and $n_{p}$ represents the depth of the prompt (the number of layers embedded in the PLM). 
The prompts are then fused with $\boldsymbol{E}^{x}$ and $\boldsymbol{E}^{C}$ respectively.
Technically, we map $\boldsymbol{P}^{x}$, $\boldsymbol{P}^{c}$, $\boldsymbol{E}^{x}$ and $\boldsymbol{E}^{C}$ to the same feature space:
\begin{equation}
\label{deqn_ex5a}
\begin{gathered}
\hat{\boldsymbol{P}}^{x}=\boldsymbol{W}^{p}\boldsymbol{P}^{x}+\boldsymbol{b}^{p}, \enspace \hat{\boldsymbol{P}}^{c}=\boldsymbol{W}^{p}\boldsymbol{P}^{c}+\boldsymbol{b}^{p}, \\
\hat{\boldsymbol{E}}^{x}=\boldsymbol{W}^{x}\boldsymbol{E}^{x}+\boldsymbol{b}^{x}, \enspace \hat{\boldsymbol{E}}^{C}=\boldsymbol{W}^{c}\boldsymbol{E}^{C}+\boldsymbol{b}^{c},
\end{gathered}
\end{equation}
where $\boldsymbol{W}^{p}$, $\boldsymbol{W}^{x}$, $\boldsymbol{W}^{c}\in\mathbb{R}^{d_{h}\times d_{h}}$ and $\boldsymbol{b}^{p}$, $\boldsymbol{b}^{x}$, $\boldsymbol{b}^{c}\in\mathbb{R}^{d_{h}}$ are trainable weights and biases.

Then $\hat{\boldsymbol{E}}^{x}$ is integrated into $\hat{\boldsymbol{P}}^{x}$ through the attention mechanism.
We calculate the attention score between any two representations in $\hat{\boldsymbol{P}}^{x}$ and $\hat{\boldsymbol{E}}^{x}$. 
For the $j$-th representation $\hat{\boldsymbol{p}}^{x}_{i,j}\in\mathbb{R}^{d_{h}}$ of the $i$-th layer in $\hat{\boldsymbol{P}}^{x}$, where $1\leqslant i\leqslant n_{p}$, $1\leqslant j\leqslant l_{p}$, and the $k$-th representation $\hat{\boldsymbol{
e}}^{x}_{k}\in\mathbb{R}^{d_{h}}$ in $\hat{\boldsymbol{E}}^{x}$, where $1 \leqslant k \leqslant n$,
we compute the dot product of $\hat{\boldsymbol{p}}^{x}_{i,j}$ and $\hat{\boldsymbol{e}}^{x}_{k}$ and obtain attention scores:
\begin{equation}
\label{deqn_ex6a}
\begin{gathered}
U_{i,j,k}^{x}=\frac{\exp(\hat{\boldsymbol{p}}^{x}_{i,j} \cdot \hat{\boldsymbol{
e}}^{x}_{k})}{\sum_{k=1}^{n}\exp(\hat{\boldsymbol{p}}^{x}_{i,j} \cdot \hat{\boldsymbol{
e}}^{x}_{k})} \,,
\end{gathered}
\end{equation}
We use the same method to integrate $\hat{\boldsymbol{E}}^{C}$ into $\hat{\boldsymbol{P}}^{c}$.
But before that, we first average $\hat{\boldsymbol{E}}^{C}$ using average pooling ($\in\mathbb{R}^{l}$) to get a sentence-level representation of the label $\hat{\boldsymbol{E}}^{c}=\text{AvgPooling}(\hat{\boldsymbol{E}}^{C})\in\mathbb{R}^{L\times d_{h}}$.
Then the subsequent operation is the same as the operation on the context, so the attention score $U_{i,j,k}^{c}$ can be obtained.

Afterwards, knowledge-enhanced context prompt $\boldsymbol{Q}^{x}\in\mathbb{R}^{l_{p}\times n_{p} \times d_{h}}$ and label prompt $\boldsymbol{Q}^{c}\in\mathbb{R}^{l_{p}\times n_{p} \times d_{h}}$ are obtained:
\begin{equation}
\label{deqn_ex9a}
\begin{gathered}
\hat{\boldsymbol{q}}^{x}_{i,j}=\hat{\boldsymbol{p}}^{x}_{i,j}+\sum_{k=1}^{n}U^{x}_{i,j,k}\cdot \hat{\boldsymbol{
e}}^{x}_{k} \,, \\
\boldsymbol{q}^{x}_{i,j}=\text{Tanh}(\hat{\boldsymbol{W}}^{x}\hat{\boldsymbol{q}}^{x}_{i,j}+\hat{\boldsymbol{b}}^{x})\,, \\
\hat{\boldsymbol{q}}^{c}_{i,j}=\hat{\boldsymbol{p}}^{c}_{i,j}+\sum_{k=1}^{L}U^{c}_{i,j,k}\cdot \hat{\boldsymbol{
e}}^{c}_{k}\,, \\
\boldsymbol{q}^{c}_{i,j}=\text{Tanh}(\hat{\boldsymbol{W}}^{c}\hat{\boldsymbol{q}}^{c}_{i,j}+\hat{\boldsymbol{b}}^{c})\,,
\end{gathered}
\end{equation}
where $\text{Tanh}(\cdot)$ is the hyperbolic tangent function, and $\hat{\boldsymbol{W}}^{x}$, $\hat{\boldsymbol{W}}^{c}\in\mathbb{R}^{d_{h}\times d_{h}}$ and $\hat{\boldsymbol{b}}^{x}$, $\hat{\boldsymbol{b}}^{c}\in\mathbb{R}^{d_{h}}$ are trainable weights and biases. $\boldsymbol{q}^{x}_{i,j}\in\mathbb{R}^{d_{h}}$ and $\boldsymbol{q}^{c}_{i,j}\in\mathbb{R}^{d_{h}}$ are the $j$-th representations of $i$-th layer in $\boldsymbol{Q}^{x}$ and $\boldsymbol{Q}^{c}$.
Finally, concatenate the $\boldsymbol{Q}^{x}$ and $\boldsymbol{Q}^{c}$ to lead to the knowledge-enhanced deep prompt $\boldsymbol{Q}=[\boldsymbol{Q}^{x};\boldsymbol{Q}^{c}]\in\mathbb{R}^{2l_{p}\times n_{p}\times d_{h}}$.

\subsection{Layer-wise Knowledge Infusion into PLM}

We now insert the deep prompts $\boldsymbol{Q}$ into different layers of PLM for a deep interaction \cite{ptuingv2}.
We use $\boldsymbol{Q}_{i}\in\mathbb{R}^{2l_{p}\times d_{h}}$ to represent the $i$-th layer of $\boldsymbol{Q}$. 
Specifically, we concatenate $\boldsymbol{Q}_{i}$ with the output $\boldsymbol{O}_{i-1}\in\mathbb{R}^{n \times d_{h}}$ of the ($i$-1)-th layer of the PLM as the input of the $i$-th layer of the PLM. 
The input of the first layer of the PLM is the concatenation of $\boldsymbol{Q}_{1}$ and $\boldsymbol{H}^{x}$:
\begin{equation}
\label{deqn_ex10a}
\begin{gathered}
\begin{cases}\boldsymbol{O}_{1}=\text{PLM}_{1}([\boldsymbol{Q}_{1};\boldsymbol{H}^{x}])\,, i=1\,, \\
\boldsymbol{O}_{i}=\text{PLM}_{i}([\boldsymbol{Q}_{i};\boldsymbol{O}_{i-1}]), 1< i\leqslant n_{p}\,.
\end{cases}
\end{gathered}
\end{equation}

\vspace{-2mm}
\subsection{Output and Learning}

Finally, the output $\boldsymbol{O}_{n_{p}}$ of the last layer of PLM is used to predict the \emph{BIO} label
sequence via $\boldsymbol{y}=\text{Softmax}(\boldsymbol{O}_{n_{p}})$.
For each $X$, the goal of our few-shot tuning (training) is to minimize the negative log-likelihood loss between the predicted probability sequence $\boldsymbol{y}$ and the corresponding gold probability sequence $\boldsymbol{y}^{g}$,
formalized as:
\begin{equation}
\label{deqn_ex11a}
\mathcal{L}=-\frac{1}{n}\sum_{i=1}^{n}\sum_{r\in C_{o},r=1}^{L}\boldsymbol{y}^{g}_{i,r}\text{log}\boldsymbol{y}_{i,r} \,,
\end{equation}
where $\boldsymbol{y}^{g}$ is a binary vector, and $\boldsymbol{y}^{g}_{i,r}$, $\boldsymbol{y}_{i,r}$ represent the golden probability and predicted probability that the label of the $i$-th word is $r$.
$C_{o}\in\mathbb{R}^{L}$ represents the original label set. 

\section{Experimental Setups}

\subsection{Datasets}
We evaluate our methods on five public benchmark NER datasets, covering a wide range of domains. 
CoNLL-2003 \cite{conll2003} on news domain, WNUT-2017 \cite{wnut17} on social domain, MIT-Moive \cite{mitmovie} and MIT-Restaurant \cite{mitrest} on review domain, JNLPBA \cite{jnlpba} on biology domain. 
Table \ref{tab:table_dataset_statistics} shows the statistics of original datasets we
use in the main experiments.

\begin{table}[!t]
\caption{Statistics on five common benchmark datasets. 
\label{tab:table_dataset_statistics}}
\fontsize{7.5}{11.5}\selectfont
\setlength{\tabcolsep}{0.8mm}
\centering
\resizebox{0.4\textwidth}{!}{
\begin{tabular}{l c c c c}
\hline
{}&\bf{Domain}&\bf{\#Train}&\bf{\#Test}&\bf{\#Entity}\\
\hline
\bf{CoNLL-2003}&News&14,042&3,454&4\\	\bf{WNUT-2017}&Social&3,395&1,288&6\\	\bf{MIT-Movie}&Review&7,817&1,954&12\\	\bf{MIT-Restaurant}&Review&7,661&1,522&8\\	\bf{JNLPBA}&Biology&16,806&3,855&5\\	
\hline
\end{tabular}
}
\end{table}

\subsection{Baselines}
{\bf{StructShot}} \cite{struct} utilizes a nearest neighbor classifier for few-shot predictions.
{\bf{CONTaiNER}} \cite{container} leverages contrastive learning to infer the distributional distance of Gaussian embeddings of entities.
{\bf{Prototype}} \cite{prototype} builds a prototypical network \cite{prototypenetwork} and utilizes the nearest neighbor criterion to assign the entity category. 
{\bf{LabelBERT}} \cite{labelbert} uses the semantics of label names as additional signals and priors. 
{\bf{EntLM}} \cite{entlmner} is a few-shot NER method that leverages an entity-oriented LM objective. 
{\bf{TemplateNER}} \cite{template} is a prompt-based approach, which enumerates all possible n-gram spans as templates and classifies each of them. 
{\bf{LightNER}} \cite{lightner} construct deep prompts into self-attention layers via a bootstrap mechanism. 
{\bf{SEE-Few}} \cite{seefew} uses context clues and entity type information to recombine candidate spans into entities and then classify them.

By the time we submit this paper, there are also some other few-shot NER works.
However, we do not make comparisons with some of them for some reasons.

\paragraph{Unfair Comparison}
We do not compare with the methods of f Chen et al. (2022) \cite{chen-etal-2022-shot} and Ji et al. (2022) \cite{entity-base}, because the methods are pre-training on a large corpus, but our method does not require such a pre-training stage.

\paragraph{Different Data}
Although Huang et al. (2022) \cite{copner} report the model performances without the pre-training stage in the original paper, they only have the results of the CoNLL-2003 and MIT-Movie datasets, where the MIT-Moive (with more sentences and entities) used by  Huang et al. (2022) \cite{copner} is different from ours. For the CoNLL-2003 dataset, our method performs better in all three settings.

\paragraph{Inaccessible Code}
In addition, , Ji et al. (2022) \cite{entity-base} do not disclose their source code.
Thus we are unable to make comparisons under our settings.

\subsection{Implementation Details}
\label{sec:Implementation Details}

\begin{table}[!t]
\caption{Hyper-parameter settings.
\label{tab:table_Hyper-parameter}}
\fontsize{7.5}{11.5}\small
\setlength{\tabcolsep}{0.8mm}
\centering
\resizebox{0.35\textwidth}{!}{
\begin{tabular}{c c}
\hline
\bf Hyper-parameter & \bf value\\
\hline
Epoch & 200\\
$l_{p}$ & 12, 24, 36, 48, 60\\
Batch Size & 16\\
Learning Rate & 5e-5\\
$n$&128\\
$l$&10\\
$d_{h}$&768\\
$n_{p}$&12\\
Dropout & 0.1\\
\hline
\end{tabular}
}
\end{table}
We conduct experiments on 5-shot, 10-shot and 20-shot respectively. For the k-shot setting, we sample k instances of each entity type from the training set. We randomly sample five different training sets and one development set for each k, each training sample tuned at different prompt lengths, then we report the mean and standard deviation of the five training samples.

We evaluate our model using precision (P), recall (R) and F1, where a predicted entity is considered correct when its sequence and type are exactly the same as the golden entity. 
In addition, the effects of all baseline models are obtained without using source domain data. We also reproduce the effects of datasets not reported in baseline articles. Except that TemplateNER is based on BART-base \cite{PLM_bart}, other models are based on BERT-base \cite{PLM_BERT}.
During training, only the knowledge-enriched deep prompt building in Figure \ref{model_frame} is trainable, and the other parts are frozen. The hyperparameter settings of our model are shown in Table \ref{tab:table_Hyper-parameter}. Our model is implemented using PyTorch and trained using NVIDIA RTX 3090 GPU.


\begin{table*}[!t]
\caption{
Comparison of our method with baseline models. Values in parentheses are standard deviations, \textcolor{bc}{\textbf{green score}} is the highest result in each column on the dataset, \textcolor{uc}{blue score} is the highest result in each column on the dataset Two high results.
\label{tab:main_resule}}
\fontsize{9.5}{11.5}\selectfont
\setlength{\tabcolsep}{1.mm}
\centering
\resizebox{1\textwidth}{!}{
\begin{tabular}{l l c c c c c c c c}
\multicolumn{2}{c}{}&\phantom{}& \textbf{\texttt{5-Shot}}&\textbf{\texttt{10-Shot}}&\textbf{\texttt{20-Shot}}&\phantom{}&\textbf{\texttt{5-Shot}}&\textbf{\texttt{10-Shot}}&\textbf{\texttt{20-Shot}}\\
\hline
\multicolumn{2}{c}{\multirow{1}{*}{{\bf{}}}}&\phantom{}& \multicolumn{3}{c}{\cellcolor{backgroud} {\emph{\textbf{CoNLL-2003}}} (News)} &\phantom{}&\multicolumn{3}{c}{{\emph{\textbf{WNUT-2017}}} (Social)}\\

\multirow{2}{*}{\bf{$\bullet$ Word-semantics-based}}&StructShot&\phantom{}&\cellcolor{backgroud}45.82\tiny{(10.30)}&\cellcolor{backgroud}62.37\tiny{(10.96)}&\cellcolor{backgroud}69.51\tiny{(6.46)}&\phantom{}&20.99\tiny{(6.83)}&27.54\tiny{(4.57)}&33.13\tiny{(2.68)}\\
{}&CONTaiNER&\phantom{}&\cellcolor{backgroud}51.70\tiny{(9.97)}&\cellcolor{backgroud}63.50\tiny{(9.65)}&\cellcolor{backgroud}\textcolor{uc}{71.00}\tiny{(6.44)}&\phantom{}&21.56\tiny{(5.44)}&28.48\tiny{(4.06)}&33.85\tiny{(2.74)}\\
\arrayrulecolor{hdash}\hdashline
\multirow{2}{*}{\bf{$\bullet$ Label-semantics-based}}&	Prototype&\phantom{}&\cellcolor{backgroud}40.37\tiny{(8.06)}&\cellcolor{backgroud}52.83\tiny{(3.22)}&\cellcolor{backgroud}53.89\tiny{(1.95)}&\phantom{}&17.42\tiny{(3.67)}&20.09\tiny{(1.57)}&22.18\tiny{(0.87)}\\
{}&	LabelBERT&\phantom{}&\cellcolor{backgroud}31.78\tiny{(2.43)}&\cellcolor{backgroud}37.81\tiny{(5.64)}&\cellcolor{backgroud}51.25\tiny{(3.08)}&\phantom{}&11.47\tiny{(3.01)}&15.41\tiny{(2.79)}&23.11\tiny{(1.62)}\\
\hdashline
\multirow{3}{*}{\bf{$\bullet$ Prompt-based}}&	EntLM&\phantom{}&\cellcolor{backgroud}49.59\tiny{(8.30)}&\cellcolor{backgroud}\textcolor{uc}{64.79}\tiny{(3.86)}&\cellcolor{backgroud}69.52\tiny{(4.48)}&\phantom{}&\textcolor{uc}{24.82}\tiny{(2.90)}&\textcolor{uc}{31.28}\tiny{(1.71)}&\textcolor{uc}{34.75}\tiny{(1.61)}\\
{}&TemplateNER&\phantom{}&\cellcolor{backgroud}43.04\tiny{(6.15)}&\cellcolor{backgroud}57.86\tiny{(5.68)}&\cellcolor{backgroud}66.38\tiny{(6.09)}&\phantom{}&19.25\tiny{(2.88)}&25.53\tiny{(2.21)}&31.91\tiny{(0.99)}\\		
{}& LightNER&\phantom{}&\cellcolor{backgroud}28.62\tiny{(4.60)}&\cellcolor{backgroud}42.96\tiny{(5.47)}&\cellcolor{backgroud}65.25\tiny{(5.95)}&\phantom{}&18.13\tiny{(3.93)}&24.97\tiny{(1.55)}&31.29\tiny{(1.16)}\\
\hdashline
\multirow{1}{*}{\bf{$\bullet$  Others}}&SEE-Few&\phantom{}&\cellcolor{backgroud}\textcolor{uc}{55.21}\tiny{(3.93)}&\cellcolor{backgroud}61.99\tiny{(1.73)}&\cellcolor{backgroud}68.21\tiny{(2.60)}&\phantom{}&22.49\tiny{(4.53)}&30.69\tiny{(1.91)}&34.63\tiny{(2.59)}\\
\hdashline
\multirow{1}{*}{\bf{$\bullet$  Ours}}&TKDP&\phantom{}&\cellcolor{backgroud}\textcolor{bc}{\textbf{56.72}}\tiny{(3.54)}&\cellcolor{backgroud}\textcolor{bc}{\textbf{64.87}}\tiny{(3.27)}&\cellcolor{backgroud}\textcolor{bc}{\textbf{73.43}}\tiny{(0.69)}&\phantom{}&\textcolor{bc}{\textbf{25.94}}\tiny{(3.31)}&\textcolor{bc}{\textbf{32.48}}\tiny{(1.07)}&\textcolor{bc}{\textbf{35.69}}\tiny{(1.69)}\\

\arrayrulecolor{black}\hline
\multicolumn{2}{c}{\multirow{1}{*}{{\bf{}}}}&\phantom{}&\multicolumn{3}{c}{{\emph{\textbf{MIT-Movie}}} (Review)}&\phantom{}&\multicolumn{3}{c}{\cellcolor{backgroud}{\emph{\textbf{MIT-Restaurant}}} (Review)}\\
\multirow{2}{*}{\bf{$\bullet$ Word-semantics-based}}&StructShot&\phantom{}&41.60\tiny{(8.97)}&53.19\tiny{(5.52)}&60.42\tiny{(2.98)}&\phantom{}&\cellcolor{backgroud}31.93\tiny{(4.32)}&\cellcolor{backgroud}34.30\tiny{(2.56)}&\cellcolor{backgroud}42.69\tiny{(1.12)}\\
{}&CONTaiNER&\phantom{}&\textcolor{uc}{51.71}\tiny{(8.55)}&57.10\tiny{(5.38)}&60.65\tiny{(2.65)}&\phantom{}&\cellcolor{backgroud}\textcolor{uc}{45.50}\tiny{(8.99)}&\cellcolor{backgroud}48.81\tiny{(5.61)}&\cellcolor{backgroud}51.25\tiny{(3.14)}\\
\arrayrulecolor{hdash}\hdashline
\multirow{2}{*}{\bf{$\bullet$ Label-semantics-based}}&	Prototype&\phantom{}&31.47\tiny{(2.39)}&38.08\tiny{(1.64)}&43.64\tiny{(1.19)}&\phantom{}&\cellcolor{backgroud}44.10\tiny{(4.66)}&\cellcolor{backgroud}45.94\tiny{(3.78)}&\cellcolor{backgroud}53.46\tiny{(2.80)}\\
{}&	LabelBERT&\phantom{}&44.64\tiny{(2.89)}&53.90\tiny{(1.71)}&59.17\tiny{(1.44)}&\phantom{}&\cellcolor{backgroud}39.67\tiny{(2.54)}&\cellcolor{backgroud}48.01\tiny{(2.23)}&\cellcolor{backgroud}57.14\tiny{(1.56)}\\	
\hdashline
\multirow{3}{*}{\bf{$\bullet$ Prompt-based}}&	EntLM&\phantom{}&46.62\tiny{(9.46)}&\textcolor{uc}{57.31}\tiny{(3.72)}&\textcolor{uc}{62.36}\tiny{(4.14)}&\phantom{}&\cellcolor{backgroud}42.60\tiny{(4.26)}&\cellcolor{backgroud}\textcolor{uc}{55.90}\tiny{(1.48)}&\cellcolor{backgroud}\textcolor{bc}{\textbf{63.00}}\tiny{(1.22)}\\
{}&TemplateNER&\phantom{}&45.97\tiny{(3.86)}&49.30\tiny{(3.35)}&59.09\tiny{(0.35)}&\phantom{}&\cellcolor{backgroud}39.22\tiny{(3.26)}&\cellcolor{backgroud}46.00\tiny{(2.22)}&\cellcolor{backgroud}57.10\tiny{(0.98)}\\	
{}& LightNER&\phantom{}&26.77\tiny{(0.65)}&40.58\tiny{(0.48)}&50.61\tiny{(1.21)}&\phantom{}&\cellcolor{backgroud}30.74\tiny{(4.38)}&\cellcolor{backgroud}52.90\tiny{(2.81)}&\cellcolor{backgroud}59.97\tiny{(7.03)}\\
\hdashline
\multirow{1}{*}{\bf{$\bullet$ Others}}&SEE-Few&\phantom{}&36.81\tiny{(2.25)}&42.06\tiny{(1.08)}&50.67\tiny{(1.21)}&\phantom{}&\cellcolor{backgroud}45.25\tiny{(3.18)}&\cellcolor{backgroud}51.20\tiny{(1.48)}&\cellcolor{backgroud}60.75\tiny{(2.07)}\\
\hdashline
\multirow{1}{*}{\bf{$\bullet$ Ours}}&TKDP&\phantom{}&\textcolor{bc}{\textbf{53.46}}\tiny{(0.71)}&\textcolor{bc}{\textbf{59.56}}\tiny{(0.84)}&\textcolor{bc}{\textbf{62.87}}\tiny{(0.73)}&\phantom{}&\cellcolor{backgroud}\textcolor{bc}{\textbf{47.44}}\tiny{(3.24)}&\cellcolor{backgroud}\textcolor{bc}{\textbf{56.82}}\tiny{(0.97)}&\cellcolor{backgroud}\textcolor{uc}{61.76}\tiny{(1.67)}\\

\arrayrulecolor{black}\hline
\multicolumn{2}{c}{\multirow{1}{*}{{\bf{}}}}&\phantom{}&\multicolumn{3}{c}{\cellcolor{backgroud}{\emph{\textbf{JNLPBA}}} (Biolog)}&\phantom{}&\multicolumn{3}{c}{\emph{\textbf{Avg. of All Dataset}}}\\
\multirow{2}{*}{\bf{$\bullet$ Word-semantics-based}}&StructShot&\phantom{}&\cellcolor{backgroud}30.52\tiny{(3.14)}&\cellcolor{backgroud}35.54\tiny{(3.02)}&\cellcolor{backgroud}45.69\tiny{(2.10)}&\phantom{}&34.17& 	42.59& 	50.29 \\
{}&CONTaiNER&\phantom{}&\cellcolor{backgroud}32.77\tiny{(3.41)}&\cellcolor{backgroud}38.70\tiny{(5.61)}&\cellcolor{backgroud}46.65\tiny{(1.77)}&\phantom{}&\textcolor{uc}{40.65} &	47.32 &	52.68 \\
\arrayrulecolor{hdash}\hdashline
\multirow{2}{*}{\bf{$\bullet$ Label-semantics-based}}&	Prototype&\phantom{}&\cellcolor{backgroud}27.99\tiny{(2.28)}&\cellcolor{backgroud}36.17\tiny{(3.87)}&\cellcolor{backgroud}46.06\tiny{(1.50)}&\phantom{}&32.27 &38.62 &43.85\\
{}&	LabelBERT&\phantom{}&\cellcolor{backgroud}26.18\tiny{(3.66)}&\cellcolor{backgroud}34.67\tiny{(4.15)}&\cellcolor{backgroud}46.78\tiny{(2.16)}&\phantom{}&30.75 	&37.96& 	47.49 
\\	
\hdashline
\multirow{3}{*}{\bf{$\bullet$ Prompt-based}}&	EntLM&\phantom{}&\cellcolor{backgroud}\textcolor{bc}{\textbf{35.45}}\tiny{(2.08)}&\cellcolor{backgroud}\textcolor{uc}{41.21}\tiny{(2.19)}&\cellcolor{backgroud}\textcolor{uc}{48.03}\tiny{(1.89)}&\phantom{}&39.82 &	\textcolor{uc}{50.10} &	\textcolor{uc}{55.53} 
\\
{}&TemplateNER&\phantom{}&\cellcolor{backgroud}33.28\tiny{(2.99)}&\cellcolor{backgroud}40.44\tiny{(2.01)}&\cellcolor{backgroud}44.07\tiny{(1.54)}&\phantom{}&36.15 &	43.83 &	51.71 
\\		
{}&LightNER&\phantom{}&\cellcolor{backgroud}24.89\tiny{(1.98)}&\cellcolor{backgroud}27.41\tiny{(1.97)}&\cellcolor{backgroud}42.43\tiny{(1.20)}&\phantom{}&25.83 &	37.76 	&49.91 
\\
\hdashline
\multirow{1}{*}{\bf{$\bullet$ Others}}&SEE-Few&\phantom{}&\cellcolor{backgroud}29.08\tiny{(5.63)}&\cellcolor{backgroud}39.31\tiny{(2.12)}&\cellcolor{backgroud}46.42\tiny{(3.28)}&\phantom{}&37.77 &	45.05 &	52.14 
\\
\hdashline
\multirow{1}{*}{\bf{$\bullet$ Ours}}&TKDP&\phantom{}&\cellcolor{backgroud}\textcolor{uc}{33.57}\tiny{(1.36)}&\cellcolor{backgroud}\textcolor{bc}{\textbf{42.44}}\tiny{(1.62)}&\cellcolor{backgroud}\textcolor{bc}{\textbf{49.31}}\tiny{(1.45)}&\phantom{}&\textcolor{bc}{\textbf{43.43}} &	\textcolor{bc}{\textbf{51.23}} &	\textcolor{bc}{\textbf{56.61}}\\
\arrayrulecolor{black}\hline
\end{tabular}
}
\end{table*}
\section{Results and Analysis}
\subsection{Main Comparisons}
Table \ref{tab:main_resule} shows the overall comparison results. As can be seen from the table, our method outperforms all baseline methods in most cases. Among them, on the CoNLL-2003 data, our method wins the CONTaiNER model with at most 2.43\% (73.43-71.00) F1 on 20-shot. From the average results, our model significantly outperforms the best baseline on 5-shot, 10-shot and 20-shot, and its F1 values are improved by 2.87\%, 1.13\%, 1.08\%, respectively. In addition, we found from the average results that under more training samples, such as 10-shot and 20-shot, the prompt-based method outperforms the method based on word semantics and label semantics, while in the 5-shot setting, based on The word-semantic approach is actually better, and our model is the best in all three settings. These results demonstrate the effectiveness of our proposed method for incorporating rich knowledge into prompts.

\subsection{Ablation Experiment}
We now work on our method itself, investigating the impact of three kinds of knowledge on model performance. The comparison results are shown in the table \ref{tab:ablation_result}. From the averaged results, we see that our model performs best in all settings when combining all three types of knowledge. In particular, with all this knowledge, the average result is 7.75\% (43.43-35.68) higher (43.43-35.68) over 5-shot than F1-value of deep prompt (DPT\cite{ptuingv2}) without any knowledge, and over 10-shot An improvement of 8.03\% (51.23-43.20) and an improvement of 6.60\% (56.61-50.01) over 20-shot. Notably, our TKDP improves the DPT method on F1 by 11.53\% (25.94-11.41) on the WNUT-2017 data in the 5-shot setting. In addition, under the 10-shot and 20-shot settings of the CoNLL-2003 dataset, TKDP has improved the effect after removing context knowledge. This may be because we have integrated the entire context into the prompt without screening useful information. Thus introducing some noise.

\begin{table*}[ht]
\caption{
Ablation studies. {\rm SK (Sememe Knowledge), LK (Label Knowledge) and CK (Context Knowledge)}. Without knowledge of all three, our {\rm TKDP} is relegated to deep prompt tuning {\rm \cite{ptuingv2} (DPT}). ning framework can be broadly applied to other few-shot NLP tasks in general purpose without much effort.
All the resources of this paper will be publicly available to facilitate related research.
\label{tab:ablation_result}}
\fontsize{9.5}{11.5}\selectfont
\setlength{\tabcolsep}{0.7mm}
\centering
\resizebox{1\textwidth}{!}{
\begin{tabular}{l c c c c c c c c c c c c c}
&\phantom{}&\textbf{\texttt{5-Shot}}&\textbf{\texttt{10-Shot}}&\textbf{\texttt{20-Shot}}&\phantom{}&\textbf{\texttt{5-Shot}}&\textbf{\texttt{10-Shot}}&\textbf{\texttt{20-Shot}}&\phantom{}&\textbf{\texttt{5-Shot}}&\textbf{\texttt{10-Shot}}&\textbf{\texttt{20-Shot}}\\
\hline
\multirow{1}{*}{}&\phantom{}&\multicolumn{3}{c}{\cellcolor{backgroud}\emph{\textbf{CoNLL-2003}}}&\phantom{}&\multicolumn{3}{c}{\emph{\textbf{WNUT-2017}}}&\phantom{}&\multicolumn{3}{c}{\cellcolor{backgroud}\emph{\textbf{MIT-Movie}}}\\
\multirow{1}{*}{TKDP}&\phantom{}&\cellcolor{backgroud}{\textcolor{bc}{\bf56.72}}\tiny{(3.54)}&\cellcolor{backgroud}64.87\tiny{(3.27)}&\cellcolor{backgroud}73.43\tiny{(0.69)}&\phantom{}&{\textcolor{bc}{\bf25.94}}\tiny{(3.31)}&{\textcolor{bc}{\bf32.48}}\tiny{(1.07)}&{\textcolor{bc}{\bf35.69}}\tiny{(1.69)}&\phantom{}&\cellcolor{backgroud}{\textcolor{bc}{\bf53.46}}\tiny{(0.71)}&\cellcolor{backgroud}{\textcolor{bc}{\bf59.56}}\tiny{(0.84)}&\cellcolor{backgroud}{\textcolor{bc}{\bf62.87}}\tiny{(0.73)}\\

\multirow{1}{*}{TKDP{\scriptsize ${-}$ CK}}&\phantom{}&\cellcolor{backgroud}55.73\tiny{(4.17)}&\cellcolor{backgroud}\cellcolor{backgroud}{\textcolor{bc}{\bf66.15}}\tiny{(2.83)}&\cellcolor{backgroud}\cellcolor{backgroud}{\textcolor{bc}{\bf73.65}}\tiny{(1.57)}&\phantom{}&24.99\tiny{(3.22)}&31.01\tiny{(0.96)}&34.53\tiny{(1.90)}&\phantom{}&\cellcolor{backgroud}51.98\tiny{(0.51)}&\cellcolor{backgroud}57.37\tiny{(0.97)}&\cellcolor{backgroud}61.85\tiny{(0.67)}\\

\multirow{1}{*}{TKDP{\scriptsize ${-}$ LK}}&\phantom{}&\cellcolor{backgroud}52.90\tiny{(3.44)}&\cellcolor{backgroud}62.28\tiny{(2.48)}&\cellcolor{backgroud}68.70\tiny{(1.76)}&\phantom{}&24.41\tiny{(3.74)}&29.40\tiny{(1.65)}&34.10\tiny{(1.41)}&\phantom{}&\cellcolor{backgroud}52.33\tiny{(0.69)}&\cellcolor{backgroud}58.14\tiny{(0.67)}&\cellcolor{backgroud}61.80\tiny{(0.84)}\\

\multirow{1}{*}{TKDP{\scriptsize ${-}$ SK}}&\phantom{}&\cellcolor{backgroud}55.87\tiny{(3.64)}&\cellcolor{backgroud}63.85\tiny{(3.40)}&\cellcolor{backgroud}71.64\tiny{(1.20)}&\phantom{}&25.78\tiny{(2.97)}&32.21\tiny{(0.86)}&34.87\tiny{(2.06)}&\phantom{}&\cellcolor{backgroud}53.04\tiny{(1.43)}&\cellcolor{backgroud}58.30\tiny{(1.47)}&\cellcolor{backgroud}62.03\tiny{(0.66)}
\\

\multirow{1}{*}{TKDP{\scriptsize ${-}$ SK,LK,CK}}&\phantom{}&\cellcolor{backgroud}47.00\tiny{(1.84)}&\cellcolor{backgroud}55.22\tiny{(4.53)}&\cellcolor{backgroud}65.29\tiny{(1.01)}&\phantom{}&14.41\tiny{(4.16)}&23.43\tiny{(2.45)}&30.20\tiny{(0.78)}&\phantom{}&\cellcolor{backgroud}48.49\tiny{(1.06)}&\cellcolor{backgroud}53.94\tiny{(0.56)}&\cellcolor{backgroud}58.77\tiny{(1.16)}
\\

\hline
\multirow{1}{*}{}&\phantom{}&\multicolumn{3}{c}{\emph{\textbf{MIT-Restaurant}}}&\phantom{}&\multicolumn{3}{c}{\cellcolor{backgroud}\emph{\textbf{JNLPBA}}}&\phantom{}&\multicolumn{3}{c}{\emph{\textbf{Avg. of All Dataset}}}\\
\multirow{1}{*}{TKDP}&\phantom{}&{\textcolor{bc}{\bf47.44}}\tiny{(3.24)}&{\textcolor{bc}{\bf56.80}}\tiny{(0.97)}&{\textcolor{bc}{\textbf{61.76}}}\tiny{(1.67)}&\phantom{}&\cellcolor{backgroud}{\textcolor{bc}{\bf33.57}}\tiny{(1.36)}&\cellcolor{backgroud}{\textcolor{bc}{\bf42.44}}\tiny{(1.62)}&\cellcolor{backgroud}{\textcolor{bc}{\bf49.31}}\tiny{(1.45)}&\phantom{}&{\textcolor{bc}{\bf43.43}}&{\textcolor{bc}{\bf51.23}}&{\textcolor{bc}{\bf56.61}}\\

\multirow{1}{*}{TKDP{\scriptsize ${-}$ CK}}&\phantom{}&46.29\tiny{(2.79)}&54.54\tiny{(0.87)}&61.37\tiny{(1.29)}&\phantom{}&\cellcolor{backgroud}31.10\tiny{(2.04)}&\cellcolor{backgroud}38.73\tiny{(1.65)}&\cellcolor{backgroud}46.80\tiny{(1.91)}&\phantom{}&42.02&49.56&55.64 

\\

\multirow{1}{*}{TKDP{\scriptsize ${-}$ LK}}&\phantom{}&45.94\tiny{(2.56)}&55.14\tiny{(1.02)}&60.88\tiny{(1.35)}&\phantom{}&\cellcolor{backgroud}30.55\tiny{(2.86)}&\cellcolor{backgroud}38.67\tiny{(1.37)}&\cellcolor{backgroud}46.70\tiny{(2.03)}&\phantom{}&41.23&48.73&54.44 
 
\\

\multirow{1}{*}{TKDP{\scriptsize ${-}$ SK}}&\phantom{}&45.44\tiny{(2.97)}&54.86\tiny{(2.93)}&60.35\tiny{(1.27)}&\phantom{}&\cellcolor{backgroud}32.54\tiny{(2.13)}&\cellcolor{backgroud}42.25\tiny{(2.42)}&\cellcolor{backgroud}48.62\tiny{(1.54)}&\phantom{}&42.53&50.38&55.50 \\

\multirow{1}{*}{TKDP{\scriptsize ${-}$ SK,LK,CK}}&\phantom{}&42.16\tiny{(3.23)}&51.65\tiny{(1.01)}&57.45\tiny{(1.46)}&\phantom{}&\cellcolor{backgroud}26.36\tiny{(2.04)}&\cellcolor{backgroud}31.75\tiny{(2.74)}&\cellcolor{backgroud}38.35\tiny{(1.55)}&\phantom{}&35.68&43.20&50.01 
 \\
\hline
\end{tabular}
}
\end{table*}

\subsection{Comparison of different continuous prompt methods}
First, we briefly introduce three fused knowledge continuous prompting methods.
\begin{itemize}
     \item \textbf{Prompt-Tuning} is a method similar to Lester et al. (2021) \cite{prompt}, which inserts consecutive prompts in the input embedding sequence for tuning. In this experiment, we enrich the continuous prompt with three kinds of knowledge.
     \item \textbf{P-Tuning} is a method similar to Liu et al. (2021) \cite{ptuning}, which still inserts continuous prompts in the input embedding sequence, but it uses BiLSTM \cite{lstm} to associate prefixes and infixes. We first enrich this continuous prompt with three kinds of knowledge, and then associate the label-knowledge prompt and the context-knowledge prompt as prefixes and infixes.
     \item \textbf{Prefix-Tuning} is a deep prompt tuning method similar to Li and Liang (2021) \cite{prefix}. It uses the same prefix and infix in each layer of GPT-2 \cite{gpt2} (transformed using MLP layers). We enrich the continuous prompt with three kinds of knowledge, and then insert this continuous prompt into each layer of the pretrained model.
\end{itemize}

\begin{table*}[ht]
\caption{
Comparison of different sequential prompting modalities. The prompt embeddings of all methods in this table are combined with sememes, labels and contextual knowledge for fair comparison.
\label{tab:Different_Prompt_Tuning}}
\fontsize{9.5}{11.5}\selectfont
\setlength{\tabcolsep}{0.7mm}
\centering
\resizebox{1\textwidth}{!}{
\begin{tabular}{l c c c c c c c c c c c c c}
&\phantom{}&\textbf{\texttt{5-Shot}}&\textbf{\texttt{10-Shot}}&\textbf{\texttt{20-Shot}}&\phantom{}&\textbf{\texttt{5-Shot}}&\textbf{\texttt{10-Shot}}&\textbf{\texttt{20-Shot}}&\phantom{}&\textbf{\texttt{5-Shot}}&\textbf{\texttt{10-Shot}}&\textbf{\texttt{20-Shot}}\\
\hline
\multirow{1}{*}{}&\phantom{}&\multicolumn{3}{c}{\cellcolor{backgroud}\emph{\textbf{CoNLL-2003}}}&\phantom{}&\multicolumn{3}{c}{\emph{\textbf{WNUT-2017}}}&\phantom{}&\multicolumn{3}{c}{\cellcolor{backgroud}\emph{\textbf{MIT-Movie}}}\\
\multirow{1}{*}{\bf{$\bullet$ Ours}}&\phantom{}&\cellcolor{backgroud}{\textcolor{bc}{\bf56.72}}\tiny{(3.54)}&\cellcolor{backgroud}{\textcolor{bc}{\bf64.87}}\tiny{(3.27)}&\cellcolor{backgroud}{\textcolor{bc}{\bf73.43}}\tiny{(0.69)}&\phantom{}&{\textcolor{bc}{\bf25.94}}\tiny{(3.31)}&{\textcolor{bc}{\bf32.48}}\tiny{(1.07)}&{\textcolor{bc}{\bf35.69}}\tiny{(1.69)}&\phantom{}&\cellcolor{backgroud}{\textcolor{bc}{\bf53.46}}\tiny{(0.71)}&\cellcolor{backgroud}{\textcolor{bc}{\bf59.56}}\tiny{(0.84)}&\cellcolor{backgroud}{\textcolor{bc}{\bf62.87}}\tiny{(0.73)}\\

\multirow{1}{*}{\bf{$\bullet$ Prompt-Tuning}}&\phantom{}&\cellcolor{backgroud}40.80\tiny{(5.37)}&\cellcolor{backgroud}53.85\tiny{(0.83)}&\cellcolor{backgroud}59.95\tiny{(1.65)}&\phantom{}&18.10\tiny{(3.69)}&22.10\tiny{(2.49)}&25.91\tiny{(1.85)}&\phantom{}&\cellcolor{backgroud}45.23\tiny{(2.06)}&\cellcolor{backgroud}51.56\tiny{(1.77)}&\cellcolor{backgroud}57.28\tiny{(1.06)}\\

\multirow{1}{*}{\bf{$\bullet$ P-Tuning}}&\phantom{}&\cellcolor{backgroud}46.92\tiny{(5.23)}&\cellcolor{backgroud}56.16\tiny{(2.32)}&\cellcolor{backgroud}63.53\tiny{(1.61)}&\phantom{}&17.01\tiny{(4.22)}&23.94\tiny{(2.32)}&25.46\tiny{(2.79)}&\phantom{}&\cellcolor{backgroud}46.48\tiny{(1.31)}&\cellcolor{backgroud}51.41\tiny{(1.59)}&\cellcolor{backgroud}56.84\tiny{(1.15)}\\

\multirow{1}{*}{\bf{$\bullet$ Prefix-Tuning}}&\phantom{}&\cellcolor{backgroud}\textcolor{uc}{53.89}\tiny{(4.42))}&\cellcolor{backgroud}\textcolor{uc}{60.23}\tiny{(3.07)}&\cellcolor{backgroud}\textcolor{uc}{68.44}\tiny{(1.73)}&\phantom{}&\textcolor{uc}{23.82}\tiny{(3.53)}&\textcolor{uc}{29.38}\tiny{(2.66)}&\textcolor{uc}{33.35}\tiny{(1.79)}&\phantom{}&\cellcolor{backgroud}\textcolor{uc}{50.87}\tiny{(1.26)}&\cellcolor{backgroud}\textcolor{uc}{55.96}\tiny{(0.70)}&\cellcolor{backgroud}\textcolor{uc}{59.51}\tiny{(0.47)}
\\

\hline
\multirow{1}{*}{}&\phantom{}&\multicolumn{3}{c}{\emph{\textbf{MIT-Restaurant}}}&\phantom{}&\multicolumn{3}{c}{\cellcolor{backgroud}\emph{\textbf{JNLPBA}}}&\phantom{}&\multicolumn{3}{c}{\emph{\textbf{Avg. of All Dataset}}}\\
\multirow{1}{*}{\bf{$\bullet$ Ours}}&\phantom{}&{\textcolor{bc}{\bf47.44}}\tiny{(3.24)}&{\textcolor{bc}{\bf56.80}}\tiny{(0.97)}&{\textcolor{bc}{\textbf{61.76}}}\tiny{(1.67)}&\phantom{}&\cellcolor{backgroud}{\textcolor{bc}{\bf33.57}}\tiny{(1.36)}&\cellcolor{backgroud}{\textcolor{bc}{\bf42.44}}\tiny{(1.62)}&\cellcolor{backgroud}{\textcolor{bc}{\bf49.31}}\tiny{(1.45)}&\phantom{}&{\textcolor{bc}{\bf43.43}}&{\textcolor{bc}{\bf51.23}}&{\textcolor{bc}{\bf56.61}}\\

\multirow{1}{*}{\bf{$\bullet$ Prompt-Tuning}}&\phantom{}&40.27\tiny{(3.02)}&47.85\tiny{(3.92)}&52.73\tiny{(1.71)}&\phantom{}&\cellcolor{backgroud}25.86\tiny{(3.15)}&\cellcolor{backgroud}28.90\tiny{(2.33)}&\cellcolor{backgroud}38.21\tiny{(2.83)}&\phantom{}&34.05 &	40.85 &	46.82 
\\
\multirow{1}{*}{\bf{$\bullet$ P-Tuning}}&\phantom{}&42.57\tiny{(1.19)}&51.32\tiny{(0.87)}&57.47\tiny{(1.12)}&\phantom{}&\cellcolor{backgroud}26.85\tiny{(2.42)}&\cellcolor{backgroud}30.75\tiny{(2.69)}&\cellcolor{backgroud}38.93\tiny{(2.99)}&\phantom{}&35.96&42.69&48.45 
\\
\multirow{1}{*}{\bf{$\bullet$ Prefix-Tuning}}&\phantom{}&\textcolor{uc}{44.11}\tiny{(3.63)}&\textcolor{uc}{52.84}\tiny{(1.82)}&\textcolor{uc}{57.69}\tiny{(1.58)}&\phantom{}&\cellcolor{backgroud}\textcolor{uc}{28.84}\tiny{(2.45)}&\cellcolor{backgroud}\textcolor{uc}{37.55}\tiny{(1.75)}&\cellcolor{backgroud}\textcolor{uc}{41.98}\tiny{(1.84)}&\phantom{}&\textcolor{uc}{40.31} 	&\textcolor{uc}{47.19}& 	\textcolor{uc}{52.19}\\
\hline
\end{tabular}
}
\end{table*}

In Table \ref{tab:Different_Prompt_Tuning}, the results of other continuous prompts compared with our method are reported, and we draw the following conclusions. 1) Our method performs best under the three settings in all datasets, because we incorporate different knowledge in each layer of PLM. Specifically, in terms of average results, our method improves by 3.12\%, 4.04\%, and 4.42\% under the three settings, respectively, compared with the suboptimal prompt-tuning method (Prefix-Tuning). 2) Prefix-Tuning is suboptimal in all cases, because it is also a deep prompt, but the knowledge contained in each layer is the same and limited. 3) The average performance of P-tuning is better than Prompt-tuning, indicating that the interaction of label knowledge and context knowledge can bring about a certain performance improvement, but compared with the performance improvement brought about by deep prompts, this is not significant.

\begin{figure}[!t]
\centering
\includegraphics[width=0.45\textwidth]{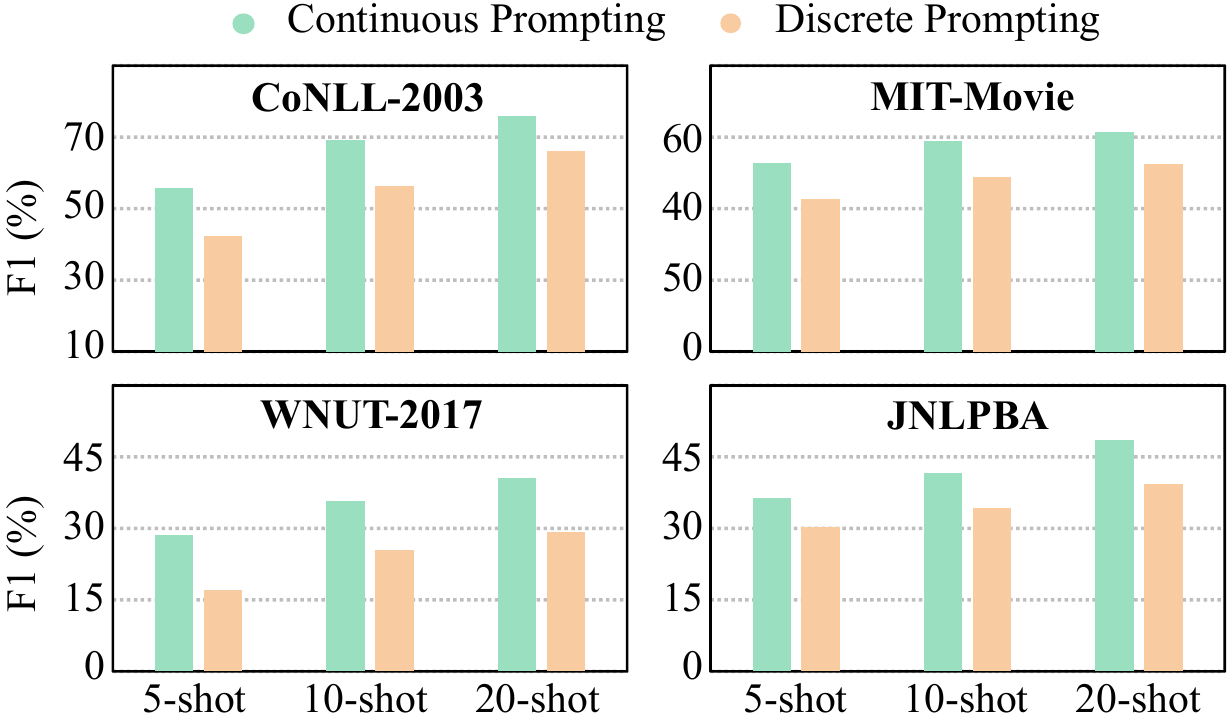}
  \caption{Contrasting methods for continuous and discrete prompting.}
  \label{Different integration methods}
\end{figure}

\subsection{Comparison of continuous and discrete prompting methods}
We skip the construction of knowledge-rich deep prompts in Figure \ref{model_frame}, and directly splice $E^x$ and $E^C$ into discrete deep prompts, which are integrated into different PLM layers for in-depth interaction. The experimental results are shown in Figure \ref{Different integration methods}. Without the prompt embedding layer, the performance of all three settings of the dataset drops significantly, because there is almost no trainable part of the whole model to fine-tune after removing the prompt embedding layer. The knowledge learned from the data is thus greatly reduced, leading to a decrease in model performance.

\subsection{Effect of prompt depth and length on model performance}
This experiment mainly explores the impact of two main parameters of prompts, namely the length of prompts and the depth of prompts, on the performance of the model. Different prompt lengths and different prompt depths will have a more significant impact on the model.

\begin{figure}[!t]
\centering
\includegraphics[width=0.45\textwidth]{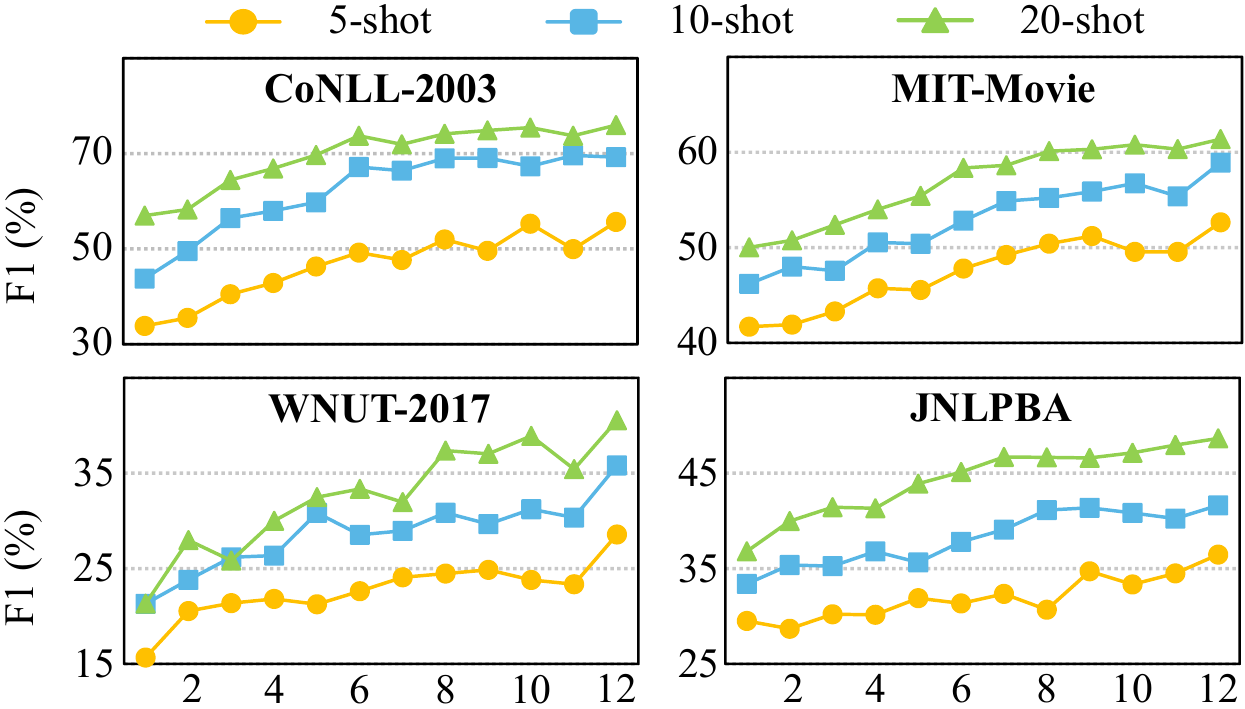}
  \caption{The effect of prompt depth $n_{p}$ on model performance.}
  \label{Different embedded depth}
\end{figure}

\begin{figure}[!t]
\centering
\includegraphics[width=0.45\textwidth]{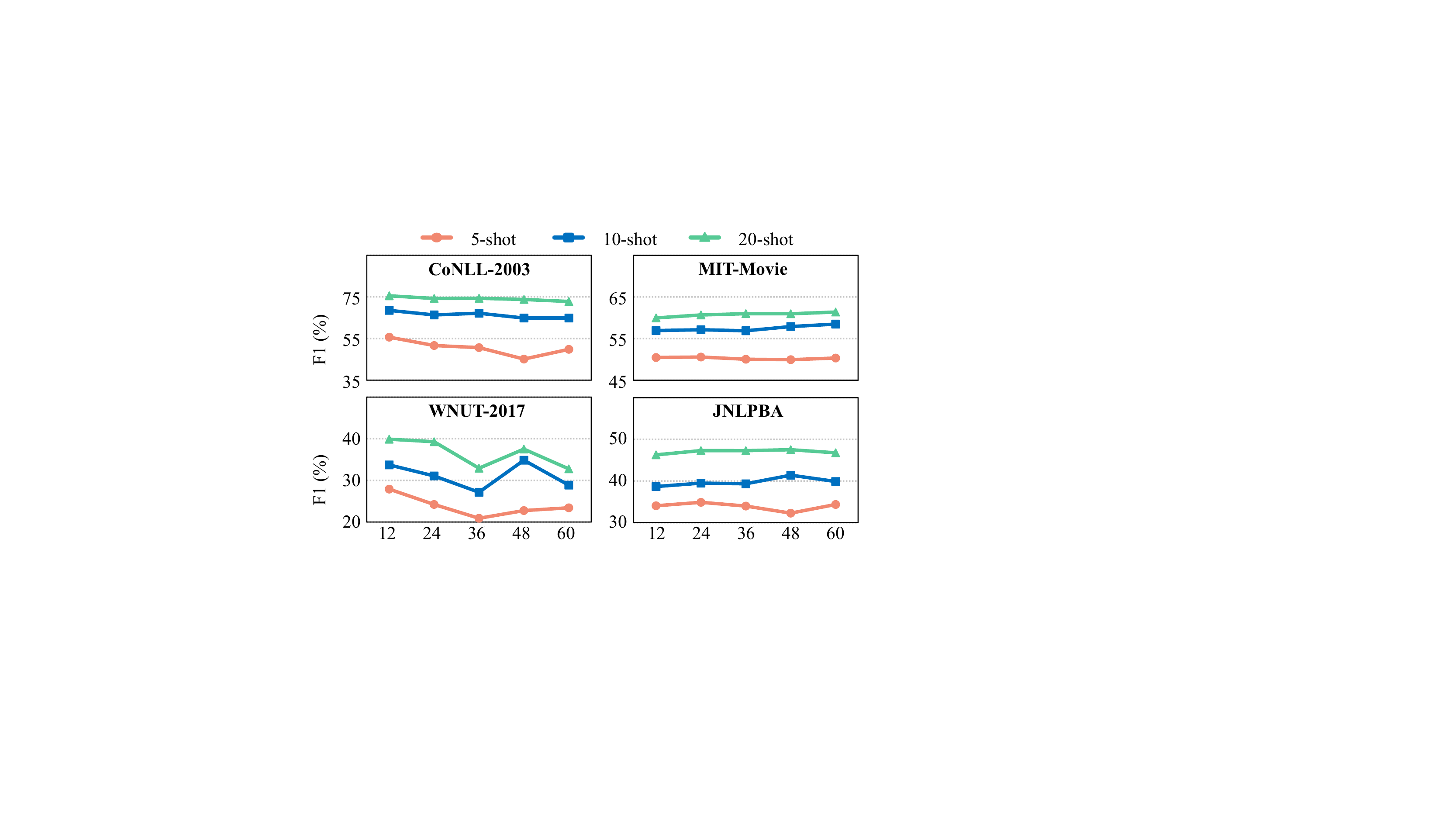}
  \caption{Effect of prompt length $l_{p}$ on model performance.}
  \label{Different prompt lengths}
\end{figure}

\begin{figure}[!t]
\centering
\includegraphics[width=0.45\textwidth]{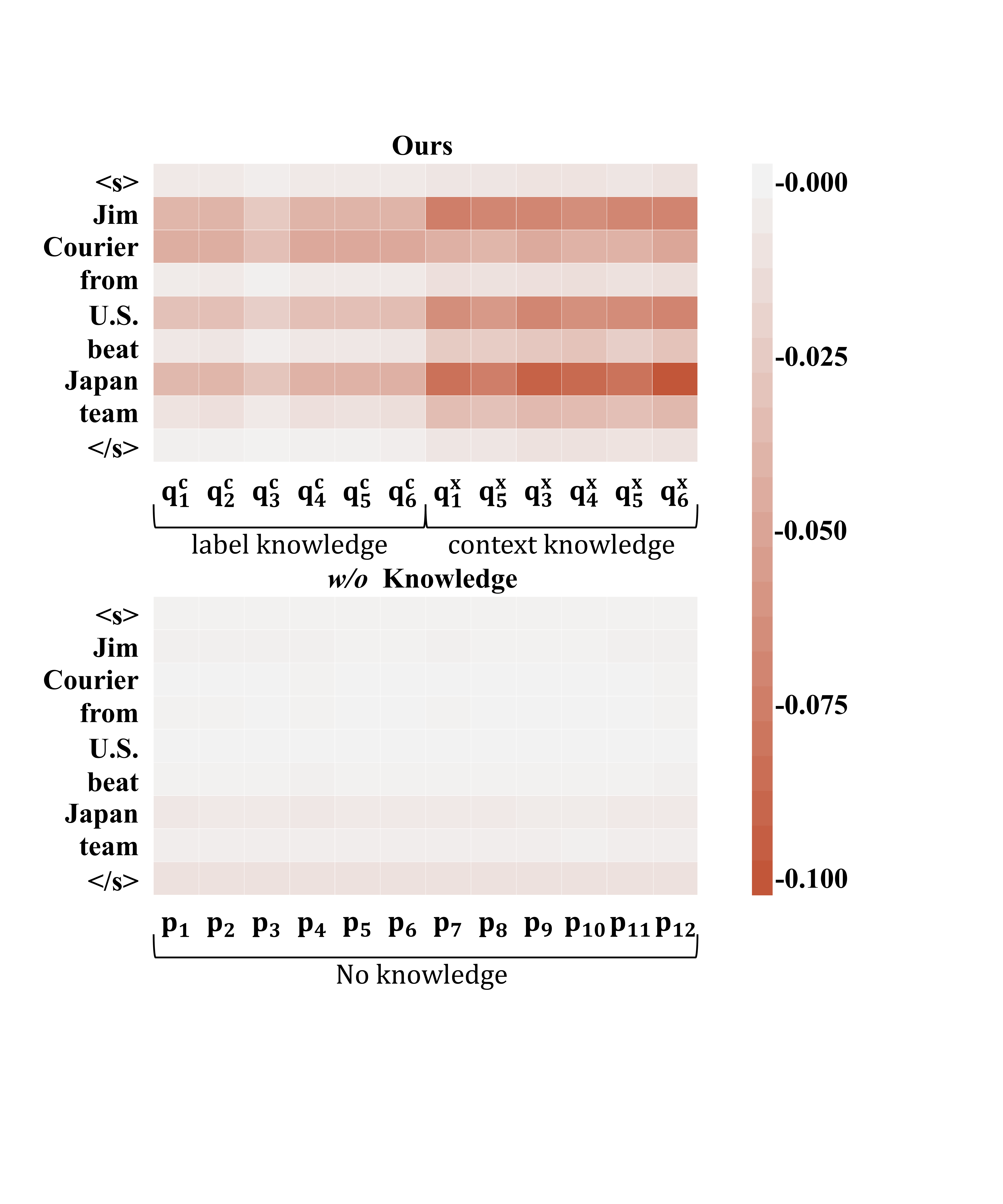}
  \caption{Attentional visualization of input pairs prompt embeddings without knowledge vs. prompt embeddings we incorporate knowledge.}
  \label{case_study}
\end{figure}

\subsubsection{Effect of prompt depth on model performance}
In Figure \ref{Different embedded depth}, we explore the effect of different depths ($n_p$) of prompt embeddings on the model, which we study on four datasets under three settings. It is clear from the trends that the model performance tends to grow with increasing embedding depth in all settings across all datasets. This is reasonable because more embedding layers bring more knowledge carrying capacity. Therefore, in our experiments, we generally set $n_p$=12.

\subsubsection{Effect of prompt length on model performance}
We also investigate the impact of different prompt lengths on model performance. In this experiment, we fix the prompt length. We still conduct experiments under three settings on four data sets, and the experimental results are shown in Figure \ref{Different prompt lengths}.

It can be seen from the figure that different datasets have different sensitivities to the prompt length under different settings. Specifically, (1) For the CoNLL-2003 and MIT-Movie datasets, the model performance does not change significantly with the change of the prompt length, but there is a slight upward trend on 20-shot and 10-shot, which may be more Longer prompts can incorporate more knowledge. (2) For the two datasets WNUT-2017 and JNLPBA, the performance of the model is greatly affected by different prompt lengths, but in general, longer prompts still have better results.

\subsection{Case study}
To help directly understand how our method manages to navigate injected knowledge and capture useful features from knowledge augmentation prompts, we end up with a case study through visualization. Figure \ref{case_study} shows the attention scores between samples in the last layer of PLM and the corresponding prompts. First, we can see that our knowledge-enhanced prompts learned to attend to different input words, suggesting that the model can indeed capture useful information. Conversely, prompts without knowledge do not provide any help for semantic reasoning. Second, we notice that the samples pay more attention to the second half of the prompt (context knowledge) rather than the first half (label knowledge). This also makes sense, since contextual features are more directly useful than label features in this case. Third, with the help of knowledge-rich prompts, entity mentions such as “Jim Courier”, “U.S” and “Japan” are highly weighted and thus help to give correct predictions.

\section{Conclusion}
We propose a threefold knowledge-enriched deep prompt tuning to achieve better few-shot NER. 
We incorporate sememe knowledge, label knowledge and context knowledge into the deep prompts. 
We conduct extensive experiments on five datasets. 
The results show that our method outperforms strong-performing baseline models on all few-shot settings.
Ablation experiments show that it is effective to integrate three types of knowledge into deep prompts, and the performance can be boosted significantly compared with the deep prompt method without knowledge. 
In addition, we found that our method consistently shows superior than the existing strong and popular prompt-based systems.
We hope this work can contribute a valuable reference for few-shot research based on prompt learning.

\section*{Acknowledgments}
This work is supported by the National Key Research and Development Program of China (No. 2022YFB3103602), the National Natural Science Foundation of China (No. 62176187), the National Key Research and Development Program of China (No. 2017YFC1200500), the Research Foundation of Ministry of Education of China (No. 18JZD015).
L Zhao would like to thank the support from Center for Artificial Intelligence (C4AI-USP), the Sao Paulo Research Foundation (FAPESP grant \#2019/07665-4), the IBM Corporation, and China Branch of BRICS Institute of Future Networks.

\bibliography{ref}
\bibliographystyle{IEEEtran}

\end{document}